\documentclass[twocolumn,final]{article}
\pdfoutput=1

\usepackage[top=2cm, left=1.5cm, bottom=2cm, right=1.5cm]{geometry}

\usepackage{amsmath}
\usepackage{amssymb}
\usepackage{latexsym}

\usepackage{graphicx}
\usepackage{natbib}
\usepackage{balance}

\usepackage{fancyhdr}
\makeatother
\usepackage{url}
\usepackage{xcolor}
\usepackage{hyperref}

\newcommand{\fscore}[1]{F\textsubscript{#1}-score}

\newcommand{\se}[1]{\textit{SE(#1)}}
\newcommand{\mset}[2]{$\mathbb{#1}^{#2}$}

\newcommand{\cellFormat}{\scriptsize \fontsize{7pt}{0pt}}
\newcommand{\shortTimes}{\! {\times} \!}

\newlength{\cellWidth}   
\newlength{\cellJump}    

\begin{document}

\twocolumn[
  \begin{@twocolumnfalse}

\title{
Roto-Translation Equivariant Convolutional Networks:\\
Application to Histopathology Image Analysis
}
\date{}

\vspace*{-50pt}
\begin{minipage}{\textwidth}
\centering
\author{
Maxime W. Lafarge\textsuperscript{ 1},
Erik J. Bekkers\textsuperscript{ 2},
Josien P.W. Pluim\textsuperscript{ 1},
Remco Duits\textsuperscript{ 2},
Mitko Veta\textsuperscript{ 1}
}
\end{minipage}

\maketitle

\vspace*{-20pt}\hspace*{40pt}\begin{minipage}{0.9\textwidth}
\footnotesize
\textsuperscript{1} \textit{Department of Biomedical Engineering, Eindhoven University of Technology, Eindhoven, The Netherlands} \newline
\textsuperscript{2} \textit{Department of Mathematics and Computer Science, Eindhoven University of Technology, Eindhoven, The Netherlands}
\end{minipage}

\vspace*{20pt}
\begin{abstract}
	Rotation-invariance is a desired property of machine-learning models for medical image analysis and in particular for computational pathology applications. 
	We propose a framework to encode the geometric structure of the special Euclidean motion group \se{2} in convolutional networks to yield translation and rotation equivariance via the introduction of \se{2}-group convolution layers. 
	This structure enables models to learn feature representations with a discretized orientation dimension that guarantees that their outputs are invariant under a discrete set of rotations.
	
	Conventional approaches for rotation invariance rely mostly on data augmentation, but this does not guarantee the robustness of the output when the input is rotated.
    At that, trained conventional CNNs may require test-time rotation augmentation to reach their full capability.
	
	This study is focused on histopathology image analysis applications for which it is desirable that the arbitrary global orientation information of the imaged tissues is not captured by the machine learning models.
	The proposed framework is evaluated on three different histopathology image analysis tasks (mitosis detection, nuclei segmentation and tumor classification).
	We present a comparative analysis for each problem and show that consistent increase of performances can be achieved when using the proposed framework.
\end{abstract}
\vspace*{20pt}

  \end{@twocolumnfalse}
]

\section{Introduction}
\label{introduction}
Invariance to irrelevant factors of variability is a desirable property of machine learning models, in particular for medical image analysis problems for which models are expected to generalize to unseen shapes, appearances, or to arbitrary orientations.
For example, histopathology image analysis problems require processing a digital slide of a stained specimen whose global orientation is strictly arbitrary.
Indeed, in the preparation workflow of histology slides, resection of the tissue is done arbitrarily and local structures within the section can have any three-dimensional orientation.
In this context, models whose output varies with the orientation of the input constitute a source of uncertainty.
The output of such image analysis systems should be rotation invariant, meaning that the output of a model should not change when its input is rotated.

Convolutional Neural Networks (CNNs) are the method of choice to solve complex image analysis tasks, in part due to the translation co-variance induced by trainable \mset{R}{2} convolution operators.
In theory, this structure allows CNNs to learn features in any orientation given sufficient capacity.
For example, if a specific edge detector is a relevant filter for the task at hand, it is expected that the CNN learns this filter in all possible directions.
Typical solutions to obtain rotation invariance consist in augmenting the dataset by generating additional randomly rotated samples, with the expectation that the model will learn the relevant features that are artificially observed under these additional orientations.
Although data augmentation is a way to induce an invariance prior, such approaches do not guarantee conventional CNNs to be rotation-invariant.
Furthermore, with such approaches it is common practice to average predictions of the trained model on a set of rotated inputs at test time: this can increase the robustness of the model, however it comes at the cost of a computational overhead.

We propose to replace convolutions in \mset{R}{2} by group convolutions using representations of the special Euclidean motion group \se{2} (roto-translation of a kernel) so as to explicitly encode the orientation of the learned features.
This structure ensures that the learned representation is co-variant/equivariant with the orientation of the input for rotations that lay on the pixel grid and to some extent for rotations that are out of the pixel grid.
We achieve orientation encoding at resolution levels higher than 90-degree via bi-linear interpolation of the \se{2} convolution kernels.
Finally rotation invariance can be achieved via a projection operation with respect to the encoded orientation of the learned representation.

\paragraph{Contributions}
This work builds upon our previous work presented at the MICCAI conference 2018 \citep{bekkers2018roto}.
In addition to a more detailed description of the proposed framework, we now present a comparative analysis of models with different angular discretization levels of the \se{2}-image representations.
Here we focus on three types of histopathology image analysis problems (mitosis detection, nuclei segmentation and tumor classification), for which we conduct experiments on popular and realistic benchmark datasets.
With this we also show that the \se{2}-image representations can be integrated in other classical CNN architectures such as U-net \citep{ronneberger2015unet}.
Finally, in a new series of in-depth experimental analyses we show an increased robustness of the proposed G-CNNs compared to standard CNNs with respect to rotational variations in the data.
This includes a quantitative and qualitative assessment of rotational invariance of the trained networks, as well as a data regime analysis in which we investigate the effect of increased angular resolution when the data availability is reduced.

\section{Rotation Invariance, Related Work, and Contributions}
\subsection{Rotation Invariance via G-CNNs}
We distinguish between invariance and equivariance/covariance as follows.
An artificial neural network (NN) is invariant with respect to certain transformations when the output of the network does not change under transformations on the input. We call a NN equivariant, or covariant\footnote{Terminology changes between fields of study (mathematics, physics, machine learning) and often refer to the same. Following custom in machine learning research we will use the term equivariance.}, when the output transforms in a predictable way when the input is transformed (we formalize this statement in Subsec.~\ref{gCNNrepresentation}). The property of equivariance guarantees that no information is lost when the input is transformed. Standard CNNs are equivariant to translations: if the input is translated the output translates accordingly and we do not need to worry about learning how to deal with translated inputs. It turns out that \emph{group convolution layers} are the only type of linear NN layers that are guaranteed to be equivariant (see e.g. \citep[Thm.~1]{bekkers_b-spline_2019}) and that the standard convolution layer is a special case that is translation equivariant. In this paper, we construct \se{2} equivariant group convolution layers and with it build G-CNNs with which we solve problems in histopathology that require rotation invariance.

Nowadays, rotation invariance is often still dealt with via data augmentations. In such an approach the data is rotated during training time while keeping the target label fixed, thereby aiming for the network to learn how to classify input samples regardless of their orientation. Downsides of this approach are that 1) valuable network capacity is spend on learning geometric behavior at the cost of descriptive representation learning, 2) rotation invariance is not guaranteed, and 3) augmentation only captures geometric invariance globally.
G-CNNs solve these problems by hard-coding geometric structure into the network architecture such that 1) geometric behavior does not have to be learned, 2) rotation invariance is guaranteed by construction, and 3) each group convolution layer achieves local equivariance on its own, so that global equivariance is still obtained when the layers are stacked.

The local-to-global equivariance property means that G-CNNs recognize both low-level features (e.g. edges), mid-level features (e.g. individual cells), and high-level features (e.g. tissue structure) independent of their orientations. In this paper we experimentally show that $SE(2)$ equivariant G-CNNs indeed solve all three aforementioned problems and that in fact the added geometric structures leads to networks that significantly outperform classical CNNs trained with data-augmentation.

\subsection{Related Work on G-CNNs}
\label{relatedWork}
\subsubsection{G-CNN Methods}
In the seminal work by \citet{cohen2016group} a framework is proposed for group equivariant CNNs. In G-CNNs, the convolution operator is redefined in terms of actions of a transformation group, and by consistent use of the group structure (rules for concatenating transformations) equivariance is ensured. They showed a significant performance gain of G-CNNs over classical CNNs, however, the practical applicability was limited to discrete transformation groups that leave the pixel grid intact (s.a. $90^\circ$ rotations and reflections). Subsequent work in the field focused on expanding the class of transformation groups that are suitable for G-CNNs by:
\begin{enumerate}
\item Working with a grid that has more symmetries than the standard Cartesian grid \citep{hoogeboom2018hexaconv}.
\item Expanding convolution kernels in a special basis, tailored to the transformation group of interest, that enables to build steerable CNNs \citep{worrall2017harmonic}
\item Relying on interpolation methods to transform kernels \cite{bekkers2018roto}, or relying on analytic basis functions and sample the transformed kernels at arbitrary resolution \citep{weiler2017learning,bekkers2018template}.
\end{enumerate}
Extensions to 3D transformation groups are described in \citep{worrall2018cubenet,winkels2019nodule,weiler20183d,andrearczyk2019pulmonary}, generalization to equivariance beyond roto-translations are described in \citep{bekkers_b-spline_2019,worrall_deep_2019}, extension to spherical data are described in  \citep{cohen2018spherical,kondor_generalization_2018,thomas_tensor_2018,esteves_learning_2018}, and additional theoretical results and further generalizations of G-CNNs are described in \citep{cohen_general_2018,kondor_generalization_2018,cohen2019gauge}.
Applications of G-CNN methods in medical image analysis are discussed below in Subsec.~\ref{sec:GCNNsInMedIA}. 

Although the first of the above generalizations elegantly enables an exact implementation of G-CNNs of roto-translations with a finer resolution than the $90^\circ$ rotation angles of \citep{cohen2016group}, it is a very specific approach that does not generalize well to other groups. The second approach does not require to sample transformed kernels at all, but works exclusively by manipulations of basis coefficients in a similar way as standard 2D convolutions (and translations) can be described in the Fourier domain. This approach however requires careful bookkeeping of the coefficients, only optimizes over kernels expressible by the basis, and the choice for non-linear activation functions is limited. In this paper we rely on the third approach. We build upon our previous work \citep{bekkers2018roto} and use bi-linear interpolation to efficiently transform (unconstrained) convolution kernels. This allows us to build \se{2} equivariant G-CNNs at arbitrary angular resolutions.

\subsubsection{Rotation Equivariant Machine Learning}
Prior, and in parallel, to the above discussed G-CNN methods, group convolution methods for pattern recognition have been proposed that, at the time, were not regarded as G-CNNs or not treated in the full generality of (end-to-end) deep learning. E.g., \citet{gens2014deep} redefine the convolution operator and construct sparse (approximative) group convolution layers that are used to build what they called deep symmetry networks. Scattering convolution networks, as proposed by \citet{mallat_group_2012}, involve a concatenation of separable group convolutions with well-designed hand-crafted filters followed by the modulus as activation function. Other examples are orientation score based template matching \citep{bekkers_training_2015}, cyclic symmetry networks \citep{dieleman2016exploiting}, oriented response networks \citep{zhou_oriented_2017}, and vector field networks \citep{marcos_rotation_2017}, which can all be considered instances of roto-translation equivariant G-CNNs.

Other techniques that focus on equivariance properties of CNNs work via transformations on input feature maps, rather than transformations of convolution kernels as in G-CNNs, and are closely related to spatial transformer networks \citep{jaderberg_spatial_2015}. These methods include warped CNNs \citep{henriques2017warped}, polar transformer networks \citep{esteves_polar_2018}, and equivariant transformer networks \citep{tai_equivariant_2019}. Although these methods describe elegant and efficient ways for achieving (global) equivariance, they often break translation equivariance and local symmetries as the transformations act globally on the whole inputs.

\subsubsection{Group Theory in Medical Image Analysis}
Equivariance constraints and group theory take a prominent position in the mathematical foundations of “classical” image analysis, e.g., in scale space and wavelet theory. In medical image analysis, group theoretical algorithms enable to respect natural equivariance constraints and deal with context and the complex geometries that are abundant in medical images.
Examples of group theoretical techniques, closely related to G-CNNs, are orientation score \citep{duits_image_2007,janssen_design_2018} methods such as crossing preserving vessel enhancement based on gauge theory on Lie groups \citep{franken_crossing-preserving_2009,hannink_crossing-preserving_2014,duits_locally_2016}, vessel and nerve fiber enhancement (in diffusion imaging) via group convolutions with Gaussian (derivative) kernels \citep{duits_left-invariant_2011,zhang_robust_2015,portegies_improving_2015}, and anatomical landmark recognition via group convolutions\citep{bekkers_b-spline_2019}. In other, non-convolutional methods in medical image analysis, group theory provides a powerful tool to deal with symmetries and geometric structure, such as in statistical shape atlases \citep{hefny_liver_2015}, shape matching \citep{hou_computing_2018}, registration \citep{arsigny_log-euclidean_2006,ashburner_fast_2007} and in general in statistics on non-Euclidean data structures \citep{pennec_riemannian_2019}. Following this successful line of geometry driven methods in medical image analysis, we propose in this paper to rely on G-CNNs to solve tasks in histopathology in an end-to-end learning setting.

\subsubsection{G-CNNs in Medical Image Analysis}
\label{sec:GCNNsInMedIA}
For many medical image analysis tasks, the location, reflection or orientation of objects of interest should not affect the output of the developed models.
Although typical solutions rely on data augmentation, several studies investigated G-CNNs in the context of medical image analysis to leverage this prior into building equivariant models that outperform classical CNNs.

In \citet{winkels2018nodule, winkels2019nodule, andrearczyk2019pulmonary}, G-CNNs were used to detect pulmonary nodules in CT scans.
G-CNNs were also investigated for segmentation tasks in dermoscopy images \citep{li2018dermoscopy}, retinal images \citep{bekkers2018roto} and microscopy images \citep{bekkers2018roto, chidester2019nuclear, graham2019rota}.
\citet{chidester2019microscopy} proposed a variation of G-CNNs for the classification of  sub-cellular protein localization in microscopy images.

Rotation-equivariant models have shown to be particularly efficient for problems in histopathology images, at cell level for mitosis detection \citep{bekkers2018roto}, nuclei segmentation \citep{chidester2019nuclear}, and at higher tissue levels for tumor classification in lymph node sections \citep{veeling2018rotation} and gland-lumen segmentation in colon histology images \citep{graham2019rota}.

\section{Material and Methods}

\begin{figure*}[ht!]
\begin{center}
\includegraphics[width=1.0\textwidth, trim=5pt 245pt 340pt 5pt, clip]{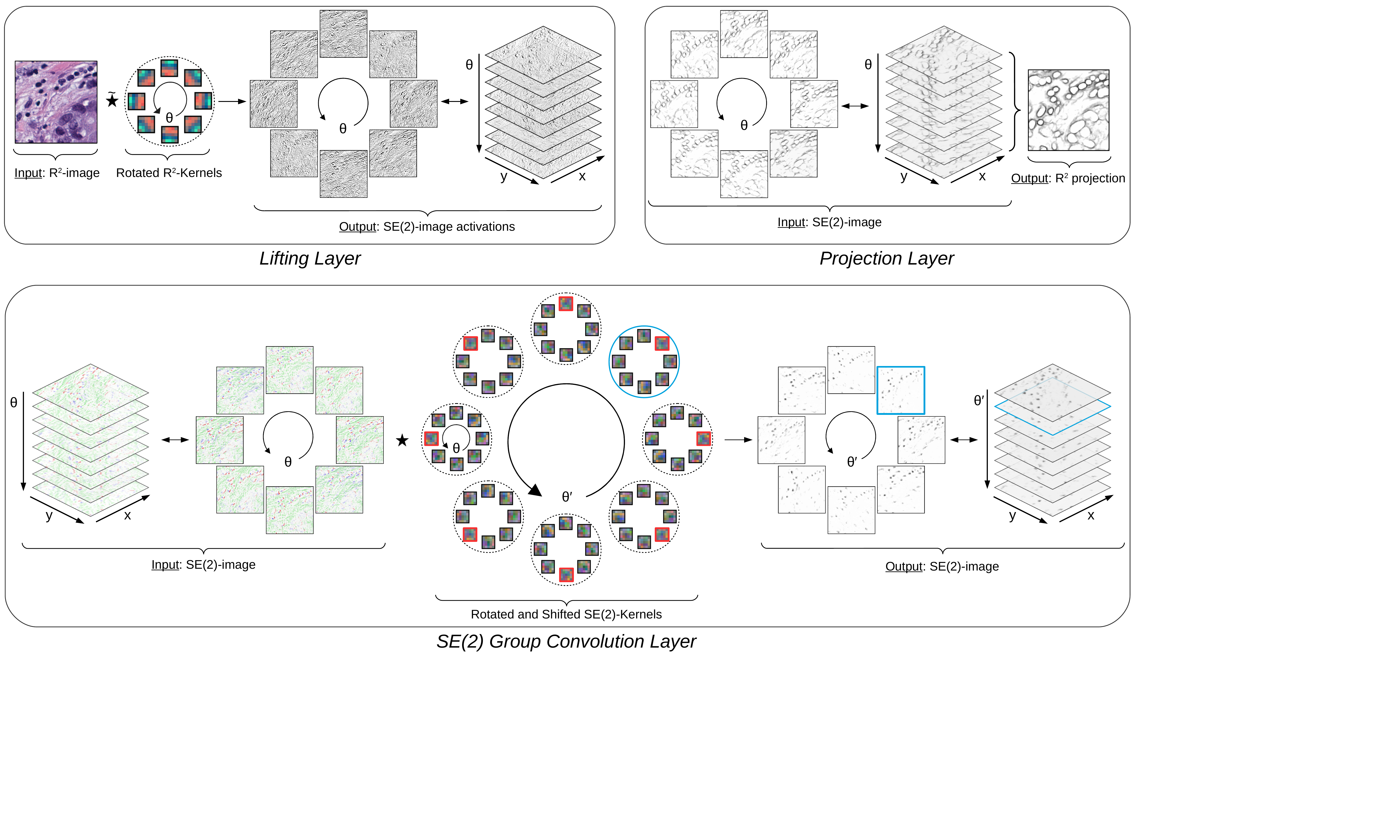}
\end{center}
\caption{
\footnotesize
Illustration of the three types of layers investigated in our G-CNNs.
The \textit{lifting layer} uses a set of rotated kernels in \mset{R}{2} to output an activation map that is an image on \se{2}.
The \textit{\se{2} group convolution layer} applies a \textit{shift-twist convolution} via a set of rotated-and-shifted kernels in \se{2} to output a \se{2}-image activation map (red border highlights the kernel transformation, cyan border highlights the output of a \se{2} kernel).
The \textit{projection layer} transforms an input \se{2}-image onto \mset{R}{2} via a rotation-invariant operation (pixel-wise maximum projection is used here).
A 3-channel input is shown for the \se{2} group convolution layer and 1-channel outputs are shown for all the layers: this is done for illustrative purposes but more channels are used in practice.
The example images used for the examples are extracted from a trained nuclei segmentation model with a 8-fold discretization of \se{2}.
}
\label{fig:mainFlowchart}
\end{figure*}

We evaluate the proposed framework on three relevant histopathology image analysis tasks: mitosis detection, nuclei classification, and patch-based tumor classification.
In this section, we first describe the benchmark datasets corresponding to the analysis tasks, that we used to train and evaluate the models.
We then describe the relationship between the proposed framework and group theory, and our proposed implementation via bi-linear interpolation of rotated convolution kernels.

\subsection{Datasets}
\label{datasets}
We chose three popular benchmark datasets of hematoxylin-eosin stained histological slides, in order to assess the performances of the proposed framework and its variants in a controlled and reproducible setup.
In these datasets, we assume that the orientation of the objects of interest is irrelevant for the classification task.

Therefore we hypothesize that any bias in the orientation information captured by a non-rotation-invariant CNN could be reflected in its performance on the selected benchmarks.
This hypothesis will be experimentally confirmed in Sect. \ref{sec:results}.

\paragraph{Mitosis Detection}
We used the public dataset \textit{AMIDA13} \citep{veta2015assessment} that consists of high power-field (HPF) images (resolution ${\sim} 0.25{\mu}\text{m}/\text{px}$) from $23$ breast cancer cases.
Eight cases (458 mitotic figures) were used to train the models and four cases (92 mitoses) for validation.
Evaluation is performed on a test set of $11$ independent cases (533 mitoses), following the evaluation procedure of the \textit{AMIDA13} challenge, for details see \citep{veta2015assessment}.

\paragraph{Multi-Organ Nuclei Segmentation}
We used the subset of the public multi-organ dataset introduced by \citep{kumar2017dataset}, that consists of $24$ HPF images (resolution ${\sim}  0.25{\mu}\text{m}/\text{px}$), selected from WSIs of four different tissue types (Breast, Liver, Kidney and Prostate), provided by \textit{The Cancer Genome Atlas} \citep{tcga2012}, associated with mask annotations of nucleus instances.
We used the balanced dataset split proposed in \citep{lafarge2019domain}: $4{\times}3$ HPF images for training (7337 nuclei), $4 {\times} 1$ HPF images for validation (1474 nuclei) and $4 {\times} 2$ HPF images for testing (4130 nuclei).
Given the high staining variability of the dataset, all the images were stain normalized using the method described in \citep{macenko2009method}.

\paragraph{Patch-Based Tumor Classification}
We used the public \textit{PCam} dataset introduced by \citep{veeling2018rotation}, that consists of $327,680$ image patches (resolution ${\sim} 1{\mu}\text{m}/\text{px}$), selected from WSIs of lymph node sections derived from the \textit{Camelyon16} Challenge \citep{bejnordi2017camelyon}.
The patches are balanced 
across the two classes (benign or malignant), based on the tumor area provided in \citep{bejnordi2017camelyon}, and we used the dataset split proposed by \citep{veeling2018rotation}.

\paragraph{Data Regime Analysis}
In order to study the behavior of the compared models when data availability is reduced, we analyzed the performances under different data regimes, by using reduced versions of the training sets.
We constructed:
\begin{itemize}
\setlength\itemsep{-0.5em}
\item Three variations of the mitosis dataset by sequentially removing two cases out of the original eight. 
\item Two variations of the nuclei dataset by sequentially removing one HPF image per organ out of the original three HPF images per organ.
\item Four variations of the patch-based tumor dataset by randomly removing $25\%$, $50\%$, $75\%$ and $90\%$ in each class-subset of the training data.
\end{itemize}

\subsection{Group Representation in CNNs}
\label{gCNNrepresentation}
\subsubsection{The Roto-Translation group \se{2}}
A group is a mathematical structure that consists of a set $G$, for example a collection of transformations, together with a binary operator $\cdot$ called the group product that satisfies four fundamental properties: \textit{Closure}: For all $h,g \in G$ we have $h\cdot g \in G$; \textit{Identiy}: There exists an identity element $e$; \textit{Inverse}: for each $g \in G$ there exists an inverse element $g^{-1} \in G$ such that $g^{-1} \cdot g = g \cdot g^{-1} = e$; and \textit{Associativity}: For each $g,h,i \in G$ we have $ (g \cdot h) \cdot i = g \cdot (h \cdot i)$. 

The group product essentially describes how two consecutive transformations, e.g. by $g,h \in G$, result in a single net transformation $(g \cdot h) \in G$. Here, we consider the group of roto-translations, denoted\footnote{It is the semi-direct product (denoted by $\rtimes$) of the group of planar translations $\mathbb{R}^2$ and rotations $SO(2)$, i.e., it is not the direct product since the rotation part acts on the translations in (\ref{eq:gprod}) in the group product of $SE(2)$.} by $SE(2) = \mathbb{R}^2 \rtimes SO(2)$, which consists of the set of all planar translations (in $\mathbb{R}^2$) and rotations (in $(SO(2)$), together with the group product given by
\begin{equation}
\label{eq:gprod}
g \cdot g' = ( \mathbf{x},\mathbf{R}_\theta ) \cdot ( \mathbf{x}', \mathbf{R}_{\theta'} )
= ( \mathbf{R}_\theta \mathbf{x}' + \mathbf{x}, \mathbf{R}_{\theta + \theta'}),\\
\end{equation}
with group elements $g = (\mathbf{x},\theta), g' = (\mathbf{x}',\theta') \in SE(2)$, with translations $\mathbf{x},\mathbf{x}'$ and planar rotations by $\theta,\theta'$. 
The group acts on the space of positions and orientations $\mathbb{R}^2 \times S^1$ via
$$
g \cdot (\mathbf{x}',\theta') = ( \mathbf{R}_\theta \mathbf{x}' + \mathbf{x}, \theta +\theta').
$$
Since $( \mathbf{x} , \mathbf{R}_\theta ) \cdot ( \mathbf{0}, 0 ) = ( \mathbf{x} , \theta )$, we can identify the group $SE(2)$ with the space of positions and orientations $\mathbb{R}^2 \times S^1$. As such we will often write $g=(\mathbf{x},\theta)$, instead of $(\mathbf{x},\mathbf{R}_\theta)$. Note that $g^{-1} = ( -\mathbf{R}_{\theta}^{-1} \mathbf{x},-\theta)$ since $g \cdot g^{-1} = g^{-1} \cdot g = (\mathbf{0},0)$.

\subsubsection{Group representations}
The structure of the group can be mapped to other mathematical objects (such as 2D images) via representations. Representations of a group $G$ are linear transformations $\mathcal{R}_g: \mathbb{L}_2(X) \rightarrow \mathbb{L}_2(X)$, parameterized by group elements $g \in G$ that transform vectors, e.g. signals/images $f \in \mathbb{L}_2(X)$ on a space $X$, and which share the group structure via
$$
(\mathcal{R}_g \circ \mathcal{R}_h)(f) = \mathcal{R}_{g \cdot h} (f), \hspace{8mm} \text{with $g,h\in G$}.
$$

We use different symbols for the representations of $SE(2)$ on different type of data structures. In particular, we write $\mathcal{R}=\mathcal{U}$ for the left-regular representation of $SE(2)$ on 2D images $f \in \mathbb{L}_2(\mathbb{R}^2)$, and it is given by
\begin{equation}
\label{eq:leftregreprSE2on2D}
(\mathcal{U}_g f) (\mathbf{x}') = f(\mathbf{R}_\theta^{-1} (\mathbf{x}' - \mathbf{x})),
\end{equation}
with $g = (\mathbf{x},\theta) \in SE(2), \; \mathbf{x}' \in \mathbb{R}^2$. It corresponds to a roto-translation of the image. We write $\mathcal{R}=\mathcal{L}$ for the left-regular representation on functions $F\in \mathbb{L}_2(SE(2))$ on $SE(2)$, which we refer to as $SE(2)$-images, and it is given by
\begin{equation}
\label{eq:leftregreprSE2}
(\mathcal{L}_g  F) (g')= F(g^{-1} \cdot g') = F(\mathbf{R}_\theta^{-1} (\mathbf{x}' - \mathbf{x}), \theta' - \theta),
\end{equation}
with $g=(\mathbf{x},\theta),g'=(\mathbf{x}',\theta') \in SE(2)$.
In Sec.~\ref{gCNNpipeline} we define the G-CNN layers in terms of these representations.

\subsubsection{Equivariance}
Given the above definitions, we can formalize the notation of equivariance. An operator $\Phi: \mathbb{L}_2(X) \rightarrow \mathbb{L}_2(Y)$ is equivariant with respect to a group $G$ if
\begin{equation}
\label{eq:equiv}
\Phi(\mathcal{R}_g(f)) = \mathcal{R}_g'(\Phi(f)),
\end{equation}
with $\mathcal{R}_g$ and $\mathcal{R}_g'$ representations of $G$ on respectively functions the domains $X$ and $Y$. I.e., if we transform the input by $\mathcal{R}_g$, then we know that the output transforms via $\mathcal{R}_g'$. To ensure that we maintain the equivariance property (\ref{eq:equiv}) of linear operators $\Phi$ it is required that we define such $\Phi$ in terms of representations of $G$, that is, via group convolutions (see e.g. \citep[Thm.~1]{bekkers_b-spline_2019}, \citep[Thm.~21]{duits_perceptual_2005}, or \citep[Thm.~6.1]{cohen_general_2018}).

\subsection{SE(2) Group Convolutional Network Layers}
\label{gCNNpipeline}

\subsubsection{Notation and 2D Convolution Layers}
In the following we denote the space of multi-channel feature maps on a domain $X$ by $(\mathbb{L}_2(X))^{N}$, with $N$ the number of channels. The feature maps themselves are denoted by $\underline{f} = (f_1,\dots,f_{N})$, with each channel $f_i \in \mathbb{L}_2(X)$. The inner product between such feature maps on $X$ is denoted by 
$$
( \underline{k} , \underline{f} )_{(\mathbb{L}_2(X))^{N}}:=\sum_{c=1}^{N}( k_c, f_c )_{\mathbb{L}_2(X)}
$$ with $( k, f )_{\mathbb{L}_2(X)}= \int_X k(\mathbf{x}') f(\mathbf{x}') {\rm d}\mathbf{x}'$ the standard inner product between real-valued functions on $X$.
Then, with these notations we note that the classical 2D cross-correlation\footnote{In CNNs one can take a convolution or a cross-correlation viewpoint and since these operators simply relate via a kernel reflection, the terminology is often used interchangeably. We take the second viewpoint, our G-CNNs are implemented using cross-correlations.} operator can defined in terms of inner products of input feature map $\underline{f}$ with translated convolution kernels $\underline{k}$ via
\begin{align}
\label{eq:2dcorr}
(\underline{k} \star_{\mathbb{R}^2} \underline{f})(\mathbf{x})
:&= (\mathcal{T}_\mathbf{x} \underline{k}, \underline{f})_{(\mathbb{L}_2(\mathbb{R}^2))^{N}} \\
&= \sum_{c=1}^{N}\int_{\mathbb{R}^2} k_c(\mathbf{x}' - \mathbf{x}) f_c(\mathbf{x}') {\rm d}\mathbf{x}',\nonumber
\end{align}
with $\mathcal{T}_\mathbf{x}$ the translation operator, the left-regular representation of the translation group $(\mathbb{R}^2,+)$. It is well known that convolution layers $\Phi$, mapping between 2D feature maps (i.e. functions on $X=Y=\mathbb{R}^2$), are equivariant with respect to translations. I.e. in Eq.~\eqref{eq:equiv} we let $\mathcal{R}_g' = \mathcal{R}_g = \mathcal{T}_\mathbf{x}$ be the left-regular representation of the translation group with $g = (\mathbf{x}) \in \mathbb{R}^2$.

\subsubsection{Roto-Translation Equivariant Convolution Layers}
Next we define two types of convolution layers that are equivariant with respect to roto-translations. We do so simply by replacing the translation operator in Eq.~\eqref{eq:2dcorr} with a representation of $SE(2)$. When the input is a 2D feature map $\underline{f} \in (\mathbb{L}_2(\mathbb{R}^2))^{N}$ we need to rely on the representation $\mathcal{U}_g$ of $SE(2)$ on 2D images, and define the \textbf{\emph{lifting correlation}}:
\begin{align}
(\underline{k}\tilde{\star} \underline{f})(g) 
:&= ( \mathcal{U}_g  \underline{k}, \underline{f} )_{(\mathbb{L}_2(\mathbb{R}^2))^{N}} \label{eq:liftingCor} \\
&= \sum_{c=1}^{N}\int_{\mathbb{R}^2} k_c(\mathbf{R}_\theta^{-1} (\mathbf{x}' - \mathbf{x})) f_c(\mathbf{x}') \, {\rm d}\mathbf{x}'. \nonumber
\end{align}
These correlations \emph{lift} 2D image data to data that lives on the 3D position orientation space $\mathbb{R}^2\times S^1 \equiv SE(2)$ by matching convolution kernels under all possible translations and rotations. 

We define the \textbf{\emph{lifting layer}}, recall Fig.~\ref{fig:mainFlowchart}, as an operator {$\tilde{\Phi}^{(l)}:(\mathbb{L}_2(\mathbb{R}^2))^{N_{l-1}}\rightarrow (\mathbb{L}_2(SE(2))^{N_{l}}$} that maps a 2D feature map $\underline{f}^{(l-1)} \in (\mathbb{L}_2(\mathbb{R}^2))^{N_{l-1}}$ with $N_{l-1}$ channels to an $SE(2)$ feature map $\underline{F}^{l} \in (\mathbb{L}_2(SE(2))^{N_{l}}$ with $N_l$ channels via lifting correlations with a collection of $N_l$ kernels, denoted with $\mathbf{k}^{(l)} := (\underline{k}_1^{(l)},\dots,\underline{k}_{N_l}^{(l)})$, each kernel with $N_{l-1}$ channels, via
\begin{equation}
\underline{F}^{(l)} = \tilde{\Phi}^{(l)}(\underline{f}^{(l-1)}) := \mathbf{k}^{(l)} \tilde{\star} \underline{f}^{(l-1)}, \label{eq:liftingLayer}
\end{equation}
where we overload the $\tilde{\star}$ symbol defined in Eq.~(\ref{eq:liftingCor}) to also denote the lifting correlation between a set of convolution kernels and a vector valued feature map via $\mathbf{k}^{(l)} \tilde{\star} \underline{f}^{(l-1)} := 
\left( \;\;
\underline{k}_{1}^{(l)} \tilde{\star} \underline{f}^{(l-1)} \;\; ,
\;\;\dots\; , \;\;
\underline{k}_{N_{l}}^{(l)} \tilde{\star} \underline{f}^{(l-1)} 
\;\;\right)$. Note that such operators are equivariant with respect to roto-translations when in \eqref{eq:equiv} we let $\mathcal{T}_g = \mathcal{U}_g$ and $\mathcal{T}_g' = \mathcal{L}_g$ be the representations of $SE(2)$ given respectively in \eqref{eq:leftregreprSE2on2D} and \eqref{eq:leftregreprSE2}, indeed $\tilde{\Phi}^{(l)}(\mathcal{U}_g \underline{f}^{(l-1)}) = \mathcal{L}_g \tilde{\Phi^{(l)}}(\underline{f}^{(l-1)})$.

The lifting layer thus generates higher-dimensional feature maps on the space of roto-translations. An $SE(2)$ equivariant layer that takes such feature maps as input is then again obtained by taking inner products of the input feature map $\underline{F}$ with (3D) roto-translated convolution kernels $\underline{K}$, where the kernels are transformed by application of the representation $\mathcal{L}_g$ of $SE(2)$ on $\mathbb{L}_2(SE(2)$).
\textbf{\emph{Group correlations}} are then defined as
\begin{align}
(\underline{K} \star \underline{F})(g) 
:&= \sum\limits_{c=1}^{N_c} ( \mathcal{L}_g K_c, F_c )_{\mathbb{L}_2(SE(2))} \label{eq:gcor}  \\
&= \sum\limits_{c=1}^{N_c} \int_{SE(2)} K_c(g^{-1} \cdot g') F_c(g') {\rm d}g'. \nonumber 
\end{align}
Note here, that a rotation of an $SE(2)$ convolution kernel is obtained via a shift-twist, a planar rotation and shift along the $\theta$-axis, see Eq.~\eqref{eq:leftregreprSE2} and Fig.~\ref{fig:mainFlowchart}. The convolution kernels $\underline{K}$ are 3-dimensional and they assign weights to activations at positions and orientations relative to a central position and orientation (relative to $g \in SE(2)$). A set of $SE(2)$ kernels $\mathbf{K}^{(l)} := (\underline{K}_1^{(l)},\dots,\underline{K}_{N_l}^{(l)})$ then defines a \textbf{\emph{group convolution layer}}, which we denote with $\Phi^{(l)}$, and which maps from $SE(2)$ feature maps $\underline{F}^{(l-1)}$ at layer $l-1$, with $N_{l-1}$ channels, to $SE(2)$-feature maps $\underline{F}^{(l)}$ at layer $l$, with $N_{l}$ channels, via
\begin{equation}\label{eq:groupConvLayer}
\underline{F}^{(l)} = \Phi^{(l)}(\underline{F}^{(l-1)}) :=
\mathbf{K}^{(l)} {\star} \underline{F}^{(l-1)},
\end{equation}
where we overload the group correlation symbol $\star$, defined in (\ref{eq:gcor}), to also denote correlation between a set of convolution kernels and a vector valued feature map on $SE(2)$ via $\mathbf{K}^{(l)} \star \underline{F}^{(l-1)} := 
\left(\;\; 
\underline{K}_{1}^{(l)} {\star} \underline{F}^{(l-1)} \;\;,\;\;
\dots \;\;,\;\;
\underline{K}_{N_{l}}^{(l)} {\star} \underline{F}^{(l-1)} 
\;\;\right).$

Finally, we define the \textbf{\emph{projection layer}} as the operator that projects a multi-channel $SE(2)$ feature map back to $\mathbb{R}^2$ via
\begin{equation}\label{eq:projectionLayer}
\underline{f}^{(l)}(\mathbf{x}) = \mathcal{P}(F^{(l)})(\mathbf{x}) := \underset{\theta \in [0,2\pi)}{\operatorname{mean}} \; \underline{F}^{(l)}(\mathbf{x},\theta).
\end{equation}
Here we define the projection layer as taking the mean over the orientation axis, however, we note that any permutation invariant operator (on the $\theta$-axis) could be used to ensure local rotation invariance, such as e.g. the commonly used $\operatorname{max}$ operator \citep{cohen2016group,bekkers2018roto}.

\subsection{Discretized \se{2,N} Group Convolutional Network}
Discretized 2D images are supported on a bounded subset of $\mathbb{Z}^2 \subset \mathbb{R}^2$ and the kernels live on a spatially rectangular grid of size $n\times n$ in $\mathbb{Z}^2$, with $n$ the kernel size.
We discretize the group $SE(2,N):=\mathbb{R}^2 \rtimes SO(2,N)$, with the space of 2D rotations in $SO(2)$ sampled with $N$ rotation angles $\theta_i{=}\frac{2\pi}{N}i$, with $i=0,\dots,N-1$. 

The discrete lifting kernels $\mathbf{k}^{(l)}$ at layer $l$, are used to map a 2D input image with $N_{l-1}$ channels to an $SE(2,N)$-image with $N_{l}$ channels, and thus have a shape of $n \times n \times N_{l-1} \times {N_{l}}$ (the discretization of $\mathbf{k}^{(l)}$ is illustrated in Fig.\ref{fig:mainFlowchart} as a set of $n$ rotated \mset{R}{2} kernels, distributed on a circle).
Likewise, the $SE(2,N)$ kernels $\mathbf{K}^{(l)}$ have a shape of $n \times n \times N \times N_{l-1} \times N_{l}$.

The lifting and group convolution layers require rotating the spatial part of the kernels and shift along the $\theta$-axis for the $SE(2)$-kernels.
We obtain the rotated spatial parts of each kernel via bi-linear interpolation.
The discretization of a single lifting kernel $k_{i,j}^{(l)}$ and its $N$ rotated versions is illustrated in the top-left part of Fig.\ref{fig:mainFlowchart}.
The discretization of a single group correlation kernel $K_{i,j}^{(l)}$ and its $N$ rotated and $\theta$-shifted versions is illustrated in the bottom part of Fig.\ref{fig:mainFlowchart}.

In order to construct the rotated sets of effective kernels $\mathbf{k}^{(l)}$ or $\mathbf{K}^{(l)}$ we rely on bi-linear interpolation. We first define a set of vectors containing base weights that are used to generate rotated versions of the same 2D kernel via bi-linear interpolation (that we implemented with a sparse matrix multiplication).
Although these sets of rotated kernels are used in the computational pipeline, only the base weights are updated during the network optimization.
By construction, the effective kernels are differentiable with respect to their base weight, enabling their update in back-propagation of gradients.

\section{Experiments}
\label{sec:experiments}
In this section, we present the G-CNN architectures that we build using the layers defined in Sec.~\ref{gCNNpipeline} and we describe the experiments that we used to analyze and validate them. In the construction of the G-CNNs we adhere to the following principle of group equivariant architecture design. 

\paragraph{G-CNN design principle} A sequence of layers starting with a lifting layer (Eq.~(\ref{eq:liftingLayer})) and followed by one or more group convolution layers (Eq.~(\ref{eq:groupConvLayer})), possibly intertwined with point-wise non-linearities, results in the encoding of roto-translation equivariant feature maps.
If such a block is followed by a projection layer (Eq.~(\ref{eq:projectionLayer})) then the entire block results in a encoding of features that is guaranteed to be rotationally invariant.
Our implementation of the G-CNN layers is available at \url{https://github.com/tueimage/se2cnn}.

\begin{table}[ht!]
\centering
\caption{
\footnotesize
Architecture of the investigated G-CNN models for mitosis detection.
The left-most column indicates the operations applied in each layer.
\textit{Max. Proj.} indicates the projection operation on \mset{R}{2}, achieved via maximum intensity projection along the orientations.
}

\begin{tabular}{p{0.18\columnwidth} || p{\cellWidth}  p{\cellWidth} p{\cellWidth} p{\cellWidth}}
\ 
& \multicolumn{4}{c}{ \small \se{2,N} Groups }
\\

\small \centering Layers
& \multicolumn{1}{c}{ \small N=1 (\mset{R}{2}) }
& \multicolumn{1}{c}{ \small N=4 (p4) }
& \multicolumn{1}{c}{ \small N=8 }
& \multicolumn{1}{c}{ \small N=16 }
\\\cline{2-5}\cline{2-5}

\scriptsize \centering Input
& \multicolumn{4}{c}{ \cellFormat $\mathsf 68 \shortTimes 68 \shortTimes 3$}
\\

 \scriptsize \centering Lifting Layer  \newline BN + ReLU  \newline MaxPool($2 \shortTimes 2$)
& \cellFormat $\mathsf 1 \shortTimes 42 \shortTimes 42 \shortTimes 16$ \newline ($1040$)
& \cellFormat $\mathsf 4 \shortTimes 42 \shortTimes 42 \shortTimes 10$ \newline ($650$)
& \cellFormat $\mathsf 8 \shortTimes 42 \shortTimes 42 \shortTimes 8$  \newline ($520$)
& \cellFormat $\mathsf 16 \shortTimes 42 \shortTimes 42 \shortTimes 6$ \newline ($390$)
\\

 \scriptsize \centering Group Conv. \newline BN + ReLU \newline MaxPool($2 \shortTimes 2$)
& \cellFormat $\mathsf 1 \shortTimes 14 \shortTimes 14 \shortTimes 16$ \newline ($5408$)
& \cellFormat $\mathsf 4 \shortTimes 14 \shortTimes 14 \shortTimes 10$ \newline ($8420$)
& \cellFormat $\mathsf 8 \shortTimes 14 \shortTimes 14 \shortTimes 8$  \newline ($10768$)
& \cellFormat $\mathsf 16 \shortTimes 14 \shortTimes 14 \shortTimes 6$ \newline ($12108$)
\\

 \scriptsize \centering Group Conv. \newline BN + ReLU \newline MaxPool($2 \shortTimes 2$)
& \cellFormat $\mathsf 1 \shortTimes 5 \shortTimes 5 \shortTimes 16$ \newline ($5408$)
& \cellFormat $\mathsf 4 \shortTimes 5 \shortTimes 5 \shortTimes 10$ \newline ($8420$)
& \cellFormat $\mathsf 8 \shortTimes 5 \shortTimes 5 \shortTimes 8$  \newline ($10768$)
& \cellFormat $\mathsf 16 \shortTimes 5 \shortTimes 5 \shortTimes 6$ \newline ($12108$)
\\

 \scriptsize \centering Group Conv. \newline BN + ReLU
& \cellFormat $\mathsf 1 \shortTimes 1 \shortTimes 1 \shortTimes 64$  \newline ($21632$)
& \cellFormat $\mathsf 4 \shortTimes 1 \shortTimes 1 \shortTimes 16$ \newline ($13472$)
& \cellFormat $\mathsf 8 \shortTimes 1 \shortTimes 1 \shortTimes 8$   \newline ($10768$)
& \cellFormat $\mathsf 16 \shortTimes 1 \shortTimes 1 \shortTimes 4$  \newline ($8072$)
\\

\scriptsize \centering Group Conv. \newline BN + ReLU
& \cellFormat $\mathsf 1 \shortTimes 1 \shortTimes 1 \shortTimes 16$  \newline ($1056$)
& \cellFormat $\mathsf 4 \shortTimes 1 \shortTimes 1 \shortTimes 16$  \newline ($1056$)
& \cellFormat $\mathsf 8 \shortTimes 1 \shortTimes 1 \shortTimes 16$  \newline ($1056$)
& \cellFormat $\mathsf 16 \shortTimes 1 \shortTimes 1 \shortTimes 16$ \newline ($1056$)
\\

\scriptsize \centering Max. Proj.
& \multicolumn{4}{c}{ \cellFormat $\mathsf 1 \shortTimes 1 \shortTimes 16$ }
\\

\scriptsize \centering FC Layer \newline Sigmoid 
& \multicolumn{4}{c}{ \cellFormat $\mathsf 1 \shortTimes 1 \shortTimes 1$ ($17$) }
\\\hline

\scriptsize \centering Total \newline Weights
& \multicolumn{1}{c}{ \scriptsize $34561$}
& \multicolumn{1}{c}{ \scriptsize $32035$ }
& \multicolumn{1}{c}{ \scriptsize $33897$ }
& \multicolumn{1}{c}{ \scriptsize $33751$ }

\end{tabular}
\label{tab:mitosisArchitecture} 
\end{table}

\subsection{Applications and Model Architectures}
\label{sec:expApplicationsAndTraining}
For each task introduced in Sect. \ref{datasets} we conducted two experiments:
first, we trained a set of variations of a baseline CNN, by changing the orientation sampling level $N$ of their \se{2,N} layers, while keeping the total number of weights of each model approximately the same.
Second, we trained each model with the reduced data regime counterparts of the training sets introduced in Sect. \ref{datasets}.

\paragraph{Mitosis Detection}
We used the mitosis classification model originally described in \cite{bekkers2018roto} as a baseline: a 6-layer CNN with three down-sampling steps, such that the overall receptive field is of size $68 \times 68$.

We designed the G-CNN variants of this baseline described in Table \ref{tab:mitosisArchitecture}, by replacing the first convolution layer by a lifting layer, replacing the following convolution layers by group convolution layers and inserting a projection layer before the last fully connected layer.

The models were trained with batches of size $64$ balanced across classes. Non-mitosis class patches were sampled based on a hard negative mining procedure \citep{cirecsan2013mitosis} using a first baseline model trained with random negative patches.
The models were trained to minimize the cross-entropy of the binary-class predictions.

\paragraph{Nuclei Segmentation}
For the nuclei segmentation task, we opted for a 7-layer U-net that corresponds to two spatial down/up-sampling operations with an overall receptive field of size $44 \times 44$.
The sequence of operations defining this G-CNN architecture is given in the first column of Table \ref{tab:nucleiArchitecture}.

The label associated with each input image is a 3-class mask corresponding to the foreground, background and border of the nuclei it contains (these masks can then be used to retrieve an individual nucleus using a segmentation procedure such as described in Sect. \ref{sec:results}).

The models were trained with batches of size $16$ balanced across patients, to minimize the class-weighted cross-entropy of the \textit{softmax} activated output maps corresponding to the three target masks.

\paragraph{Tumor Classification}
The baseline architecture we used for the tumor classification model is a 6-layer CNN with three down-sampling steps, such that the overall receptive field is of size $88 \times 88$ (see Table \ref{tab:pcamArchitecture} for the detailed architecture).

The models were trained with batches of size $64$ balanced across classes.
We refined both classes by running a hard negative mining procedure \citep{cirecsan2013mitosis} using a first baseline model trained with the original dataset of the benchmark.
The models were trained to minimize the cross-entropy of the binary-class predictions.

\begin{table}[ht!]
\centering
\caption{
\footnotesize
Architecture and weight counting of the G-CNN models for patch-based tumor classification.
The left-most column indicates the operations in each layer.
\textit{Concat(HL.$x$)} indicates the characteristic skip operation of the U-net architecture that consist in concatenating a centered crop of the output activation of the $x^{th}$ layer of the network.
\textit{Max. Proj.} indicates the projection operation on \mset{R}{2}, achieved via maximum intensity projection along the orientations.
}

\begin{tabular}{p{0.18\columnwidth} || p{\cellWidth} p{\cellWidth} p{\cellWidth} p{\cellWidth}}
\ 
& \multicolumn{4}{c}{ \small \se{2,N} Groups }
\\

\small \centering Layers
& \multicolumn{1}{c}{ \small N=1 (\mset{R}{2}) }
& \multicolumn{1}{c}{ \small N=4 (p4) }
& \multicolumn{1}{c}{ \small N=8 }
& \multicolumn{1}{c}{ \small N=16 }
\\\cline{2-5}\cline{2-5}
\scriptsize \centering Input
& \multicolumn{4}{c}{ \cellFormat $\mathsf 60 \shortTimes 60 \shortTimes 3$}
\\

 \scriptsize \centering Lifting Layer  \newline BN + ReLU \newline MaxPool($2 \shortTimes 2$)
& \cellFormat $\mathsf 1 \shortTimes 28 \shortTimes 28 \shortTimes 16$  \newline $(1040)$
& \cellFormat $\mathsf 4 \shortTimes 28 \shortTimes 28 \shortTimes 10$  \newline ($650$)
& \cellFormat $\mathsf 8 \shortTimes 28 \shortTimes 28 \shortTimes 8$  \newline ($520$)
& \cellFormat $\mathsf 16 \shortTimes 28 \shortTimes 28 \shortTimes 6$ \newline ($390$)
\\

 \scriptsize \centering Group Conv. \newline BN + ReLU \newline MaxPool($2 \shortTimes 2$)
& \cellFormat $\mathsf 1 \shortTimes 12 \shortTimes 12 \shortTimes 16$  \newline ($5408$)
& \cellFormat $\mathsf 4 \shortTimes 12 \shortTimes 12 \shortTimes 10$  \newline ($8420$)
& \cellFormat $\mathsf 8 \shortTimes 12 \shortTimes 12 \shortTimes  8$  \newline ($10768$)
& \cellFormat $\mathsf 16 \shortTimes 12 \shortTimes 12 \shortTimes 6$  \newline ($12108$)
\\

 \scriptsize \centering Group Conv. \newline BN + ReLU
& \cellFormat $\mathsf 1 \shortTimes 8 \shortTimes 8 \shortTimes 16$  \newline ($5408$)
& \cellFormat $\mathsf 4 \shortTimes 8 \shortTimes 8 \shortTimes 10$  \newline ($8420$)
& \cellFormat $\mathsf 8 \shortTimes 8 \shortTimes 8 \shortTimes  8$  \newline ($10768$)
& \cellFormat $\mathsf 16 \shortTimes 8 \shortTimes 8 \shortTimes 6$ \newline ($12108$)
\\

 \scriptsize \centering Up-sampling \newline Concat(HL.2) \newline Group Conv. \newline BN + ReLU
& \cellFormat $\mathsf 1 \shortTimes 12 \shortTimes 12 \shortTimes 16$  \newline ($10784$)
& \cellFormat $\mathsf 4 \shortTimes 12 \shortTimes 12 \shortTimes 10$  \newline ($16820$)
& \cellFormat $\mathsf 8 \shortTimes 12 \shortTimes 12 \shortTimes  8$   \newline ($21520$)
& \cellFormat $\mathsf 16 \shortTimes 12 \shortTimes 12 \shortTimes 6$  \newline ($24204$)
\\

 \scriptsize \centering Up-sampling \newline Concat(HL.1) \newline Group Conv. \newline BN + ReLU
& \cellFormat $\mathsf 1 \shortTimes 20 \shortTimes 20 \shortTimes 64$  \newline ($43136$)
& \cellFormat $\mathsf 4 \shortTimes 20 \shortTimes 20 \shortTimes 16$  \newline ($26912$)
& \cellFormat $\mathsf 8 \shortTimes 20 \shortTimes 20 \shortTimes  8$  \newline ($21520$)
& \cellFormat $\mathsf 16 \shortTimes 20 \shortTimes 20 \shortTimes 4$ \newline ($16136$)
\\

 \scriptsize \centering Group Conv. \newline BN + ReLU
& \cellFormat $\mathsf 1 \shortTimes 20 \shortTimes 20 \shortTimes  16$  \newline ($1056$)
& \cellFormat $\mathsf 4 \shortTimes 20 \shortTimes 20 \shortTimes  16$  \newline ($1056$)
& \cellFormat $\mathsf 8 \shortTimes 20 \shortTimes 20 \shortTimes  16$  \newline ($1056$)
& \cellFormat $\mathsf 16 \shortTimes 20 \shortTimes 20 \shortTimes 16$  \newline ($1056$)
\\

\scriptsize \centering Max. Proj.
& \multicolumn{4}{c}{ \scriptsize $20 \shortTimes 20 \shortTimes 16$ }
\\

\scriptsize \centering FC Layer \newline Softmax 
& \multicolumn{4}{c}{ \scriptsize $20 \shortTimes 20 \shortTimes 3$ ($54$) }
\\\hline

\scriptsize \centering Total \newline Weights
& \multicolumn{1}{c}{ \scriptsize $66886$}
& \multicolumn{1}{c}{ \scriptsize $62332$ }
& \multicolumn{1}{c}{ \scriptsize $66206$ }
& \multicolumn{1}{c}{ \scriptsize $66056$ }

\end{tabular}
\label{tab:nucleiArchitecture} 
\end{table}

\begin{table}[ht!]
\centering
\caption{
\footnotesize
Architecture and weight counting of the G-CNN models for patch-based tumor classification.
The left-most column indicates the operations in each layer.
\textit{Mean. Proj.} indicates the projection operation on \mset{R}{2}, achieved via mean intensity projection along the orientations.}

\begin{tabular}{p{0.18\columnwidth} || p{\cellWidth}  p{\cellWidth} p{\cellWidth} p{\cellWidth}}
\ 
& \multicolumn{4}{c}{ \small \se{2,N} Groups }
\\

\small \centering Layers
& \multicolumn{1}{c}{ \small N=1 (\mset{R}{2}) }
& \multicolumn{1}{c}{ \small N=4 (p4) }
& \multicolumn{1}{c}{ \small N=8 }
& \multicolumn{1}{c}{ \small N=16 }
\\\cline{2-5}\cline{2-5}
\scriptsize \centering Input
& \multicolumn{4}{c}{ \cellFormat $\mathsf 88 \shortTimes 88 \shortTimes 3$}
\\

 \scriptsize \centering Lifting Layer  \newline BN + ReLU \newline MaxPool($2 \shortTimes 2$)
& \cellFormat $\mathsf 1 \shortTimes 42 \shortTimes 42 \shortTimes 32$  \newline ($2080$)
& \cellFormat $\mathsf 4 \shortTimes 42 \shortTimes 42 \shortTimes 19$  \newline ($1235$)
& \cellFormat $\mathsf 8 \shortTimes 42 \shortTimes 42 \shortTimes 14$  \newline ($910$)
& \cellFormat $\mathsf 16 \shortTimes 42 \shortTimes 42 \shortTimes 10$ \newline ($650$)
\\

 \scriptsize \centering Group Conv. \newline BN + ReLU \newline MaxPool($2 \shortTimes 2$)
& \cellFormat $\mathsf 1 \shortTimes 19 \shortTimes 19 \shortTimes 32$  \newline ($21568$)
& \cellFormat $\mathsf 4 \shortTimes 19 \shortTimes 19 \shortTimes 19$  \newline ($30362$)
& \cellFormat $\mathsf 8 \shortTimes 19 \shortTimes 19 \shortTimes 14$  \newline ($32956$)
& \cellFormat $\mathsf 16 \shortTimes 19 \shortTimes 19 \shortTimes 10$ \newline ($33620$)
\\

 \scriptsize \centering Group Conv. \newline BN + ReLU \newline MaxPool($3 \shortTimes 3$)
& \cellFormat $\mathsf 1 \shortTimes 5 \shortTimes 5 \shortTimes 32$  \newline ($21568$)
& \cellFormat $\mathsf 4 \shortTimes 5 \shortTimes 5 \shortTimes 19$  \newline ($30362$)
& \cellFormat $\mathsf 8 \shortTimes 5 \shortTimes 5 \shortTimes 14$  \newline ($32956$)
& \cellFormat $\mathsf 16 \shortTimes 5 \shortTimes 5 \shortTimes 10$ \newline ($33620$)
\\

 \scriptsize \centering Group Conv. \newline BN + ReLU
& \cellFormat $\mathsf 1 \shortTimes 1 \shortTimes 1 \shortTimes 64$  \newline ($43136$)
& \cellFormat $\mathsf 4 \shortTimes 1 \shortTimes 1 \shortTimes 16$  \newline ($25568$)
& \cellFormat $\mathsf 8 \shortTimes 1 \shortTimes 1 \shortTimes 8$   \newline ($18832$)
& \cellFormat $\mathsf 16 \shortTimes 1 \shortTimes 1 \shortTimes 4$  \newline ($13448$)
\\

\scriptsize \centering Group Conv. \newline BN + ReLU
& \cellFormat $\mathsf 1 \shortTimes 1 \shortTimes 1 \shortTimes 16$  \newline ($1056$)
& \cellFormat $\mathsf 4 \shortTimes 1 \shortTimes 1 \shortTimes 16$  \newline ($1056$)
& \cellFormat $\mathsf 8 \shortTimes 1 \shortTimes 1 \shortTimes 16$  \newline ($1056$)
& \cellFormat $\mathsf 16 \shortTimes 1 \shortTimes 1 \shortTimes 16$ \newline ($1056$)
\\

\scriptsize \centering Mean Proj.
& \multicolumn{4}{c}{ \cellFormat $\mathsf 1 \shortTimes 1 \shortTimes 16$ }
\\

\scriptsize \centering FC Layer \newline Sigmoid 
& \multicolumn{4}{c}{ \cellFormat $\mathsf 1 \shortTimes 1 \shortTimes 1$ ($17$) }
\\\hline

\scriptsize \centering Total \newline Weights
& \multicolumn{1}{c}{ \scriptsize $89425$}
& \multicolumn{1}{c}{ \scriptsize $88600$ }
& \multicolumn{1}{c}{ \scriptsize $86727$ }
& \multicolumn{1}{c}{ \scriptsize $82411$ }

\end{tabular}
\label{tab:pcamArchitecture} 
\end{table}

\subsection{Implementation details}
For all three baseline architectures, convolution kernels are of size $5 \times 5$ with circular masking and fully connected layers are implemented as convolutional layers with kernels of shape $1 \times 1$ to enable dense application (the resulting models can efficiently be applied on larger input sizes).

Batch Normalization \citep{ioffe2015batch} is used throughout the networks.
Batch statistics are normally computed across batch and spatial dimensions of the activations, but we also included the orientation-axis of the \se{2,N}-image activation maps in the statistic computation to ensure their invariance with respect to the orientation of the input.

All models were trained with Stochastic Gradient Descent with momentum (learning rate $0.01$, momentum $0.9$) and a epoch-wise learning rate decay using a factor of $0.5$ was applied.
Training was stopped after convergence of the loss computed on the validation sets.
All models were regularized with decoupled weight decay (coefficient $5 \times 10 ^{-4}$). Baseline augmentation transformations were applied to the training image patches (random spatial transposition, random 90-degree-wise rotation, random channel-wise brightness shifting).

\subsection{Experiment: Orientation Sampling}
\label{sec:expOrientation}
In order to assess the effect of using the proposed \se{2,N} G-CNN structure on the benchmark performances, we trained every model with $N \in \{1,4,8,16\}$.
In order to allow fair comparison we adjusted the number of channels in every layer involving \se{2,N}-image representation such that the total number of weights in the models stay close to the count of the corresponding baselines.
The detailed distributions of the weights are shown in Tables \ref{tab:mitosisArchitecture}, \ref{tab:nucleiArchitecture} and \ref{tab:pcamArchitecture}: for each \se{2,N} group, the dimensions of the output of the layers are shown with the format $N {\times} Height {\times} Width {\times} C$, with $C$ the number of output channels in the layer.

Each model was trained three times with random initialization seeds.
We report the mean and standard deviation of the performances across three random intializations.

\subsection{Experiment: Data Regime Experiments}
\label{expDataRegime}
In order to assess the effect of using the proposed \se{2,N} with varying sampling factor N when data is availability is reduced, we trained each model on the data-regime subsets presented in Sect. \ref{datasets}.
Likewise, each model was trained three times with random initialization seeds so as to report the variability of the performances.

\begin{figure*}[ht!]
\begin{center}
\includegraphics[width=\textwidth, trim=5pt 415pt 55pt 5pt, clip]{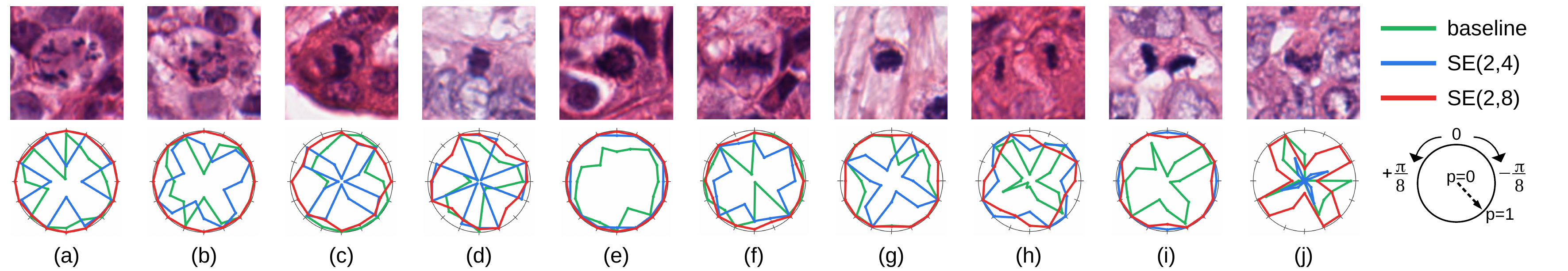}
\caption{
\footnotesize
Example of mitosis-centered image patches selected from the test set.
Below each, polar plots show model predictions (distance from origin) as a function of the orientation of the input (angle coordinate) using steps of $\pi / 8$ rad.
An ideal model would then produce a circle with maximum radius.
Selected models are indicated with colors, and correspond to the best obtained models that were trained without reduced data regime over repeats (based on their \fscore{1}).
}
\label{fig:resultsQualitativeMitosis}
\end{center}
\end{figure*}

\begin{figure*}[ht!]
\begin{center}
\includegraphics[width=\textwidth, trim=5pt 438pt 55pt 5pt, clip]{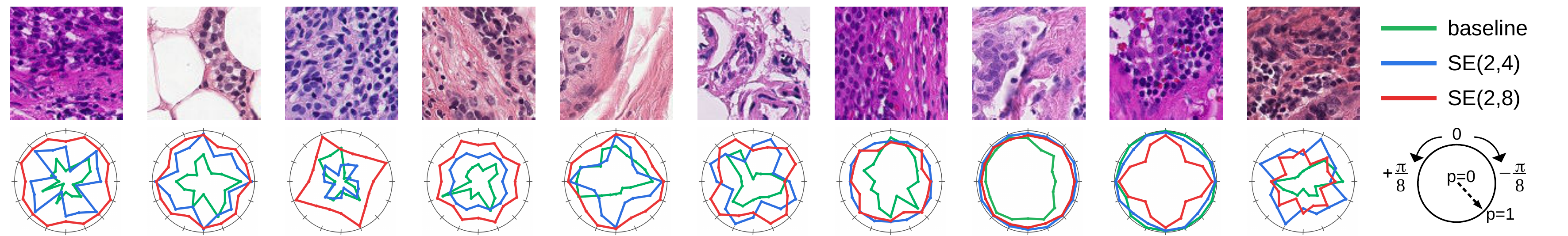}
\caption{
\footnotesize
Example of image patches selected from the test set of the \textit{PCam} benchmark, for which pixels in the center area were classified as \textit{tumor tissue}.
Below each, polar plots show model predictions (distance from origin) as a function of the orientation of the input (angle coordinate) using steps of $\pi / 8$ rad.
Selected models are indicated with colors, and correspond to the best obtained models that were trained without reduced data regime over repeats (based on their accuracy).
}
\label{fig:resultsQualitativePCam}
\end{center}
\end{figure*}

\begin{figure*}[ht!]
\begin{center}
\includegraphics[width=0.91\textwidth, trim=5pt 50pt 115pt 5pt, clip]{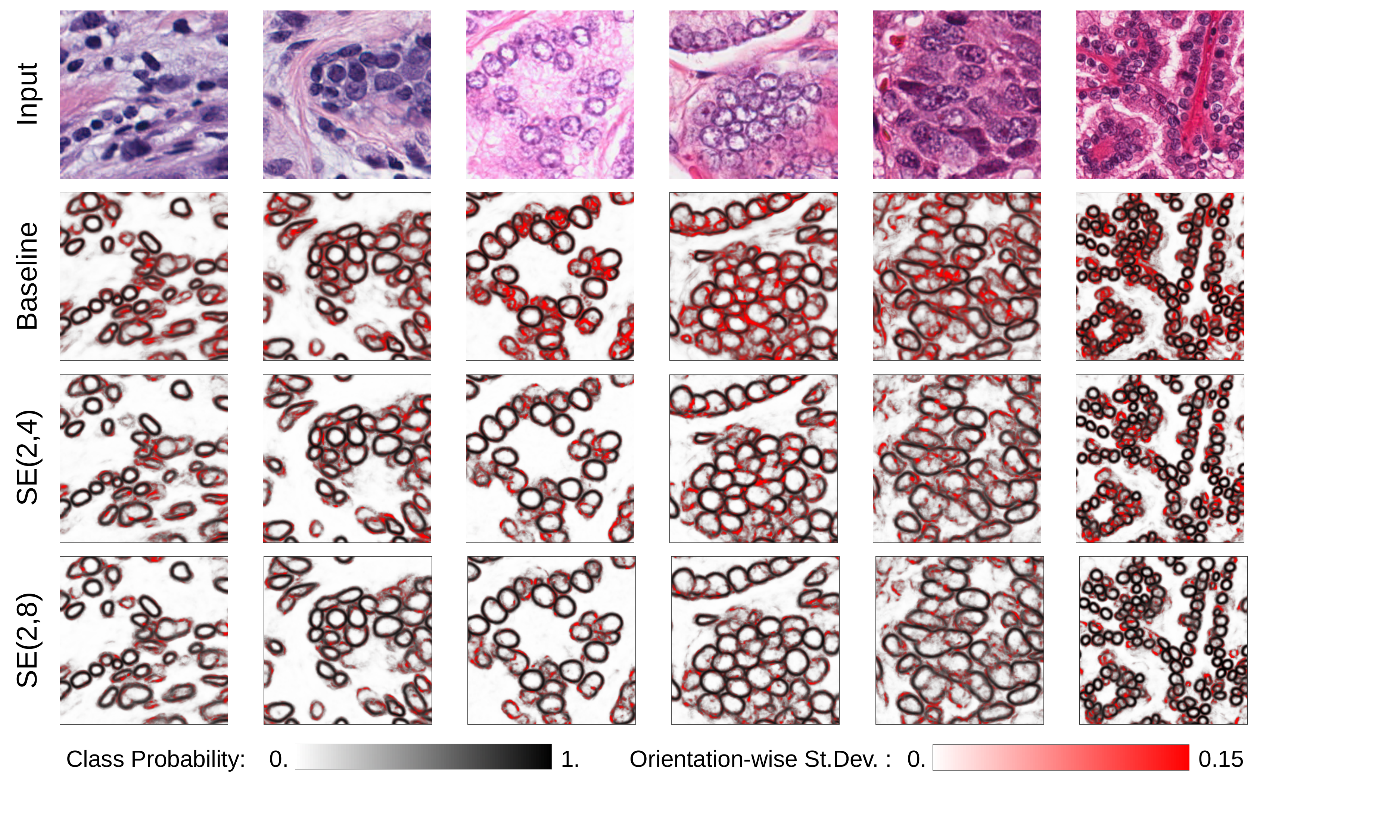}
\caption{\footnotesize
Example of image patches selected from the test set of the nuclei segmentation benchmark (column 1-2: breast tissue, column 3-4: prostate tissue, column 5: kidney tissue, column 6: liver).
For each image, and a selection of models, the raw predictions of the nucleus boundary class were computed and stored for the set of rotated inputs using steps of $\pi / 8$ rad.
Predictions were re-aligned and their means were mapped to gray-scale and the standard deviations of the predictions were mapped to a white-to-red color scale.
The overlap of these statistics is shown below each original image.
Selected models are the best obtained models that were trained without reduced data regime over repeats (based on their \fscore{1}).
}
\label{fig:resultsQualitativeNuclei}
\end{center}
\end{figure*}

\section{Results}
\label{sec:results}
This section summarizes the qualitative and quantitative results of the experiments we conducted.
Each trained model was evaluated on the test set of its corresponding benchmark dataset based on standard performance metrics.

\paragraph{Mitosis Detection}
For the mitosis detection task, models were densely applied on test images, followed by a smoothing operation before extracting all local maxima to be considered candidate detections.
We computed the \fscore{1} of the set of detections using an operating point that is optimized on the validation set, as described in the scoring protocol used in \citep{veta2015assessment}.

\paragraph{Nuclei Segmentation}
To quantify the performances of the nuclei segmentation model, generation of segmented candidate objects is obtained by following the protocol used in \citep{kumar2017dataset, lafarge2019domain}.
First, marker seeds are derived from thresholded foreground and background predictions, border predictions are used as the watershed energy landscape.
Then, candidate objects that overlap the nuclei ground-truth masks by at least 50\% of their area are considered hits, enabling object-level detection quantification to be calculated using the \fscore{1}.
Thresholds to generate marker seeds were selected such that the \fscore{1} is maximized on the validation set.

\paragraph{Patch-based tumor classification}
To evaluate the tumor classification model, we computed the class probability of every patch of the test dataset and calculated the accuracy of the model given the ground-truth labels as in \citet{veeling2018rotation} after selection of the operating point that maximizes the accuracy on the validation set.

\subsection{Qualitative Results}
\label{qualitativeResults}
We qualitatively investigated the robustness of the prediction of different models to controlled rotations of the input.
We see that the model predictions can be very inconsistent for our best baseline model, in comparison to G-CNN models (see Fig. \ref{fig:resultsQualitativeMitosis}, Fig. \ref{fig:resultsQualitativeNuclei} and \ref{fig:resultsQualitativePCam}) in particular for cell or tissue morphologies that are typically asymmetric.
For example, the mitotic figures (h) and (i) shown in Fig. \ref{fig:resultsQualitativeMitosis} are in telophase (directed separation of the pair of chromosomes) and the variance of the prediction of the baseline model is higher for these cases (green curve) compared to the G-CNN models (blue and red curves).
We also observe that for the \se{2,4} model, predictions that are obtained for an input image rotated with an angle below $\pi / 2$rad also produce some variance, but present a $\pi / 2$rad-period cyclic pattern.

\begin{figure}[ht!]
\begin{center}
\includegraphics[width=\columnwidth, trim=20pt 90pt 50pt 90pt, clip]{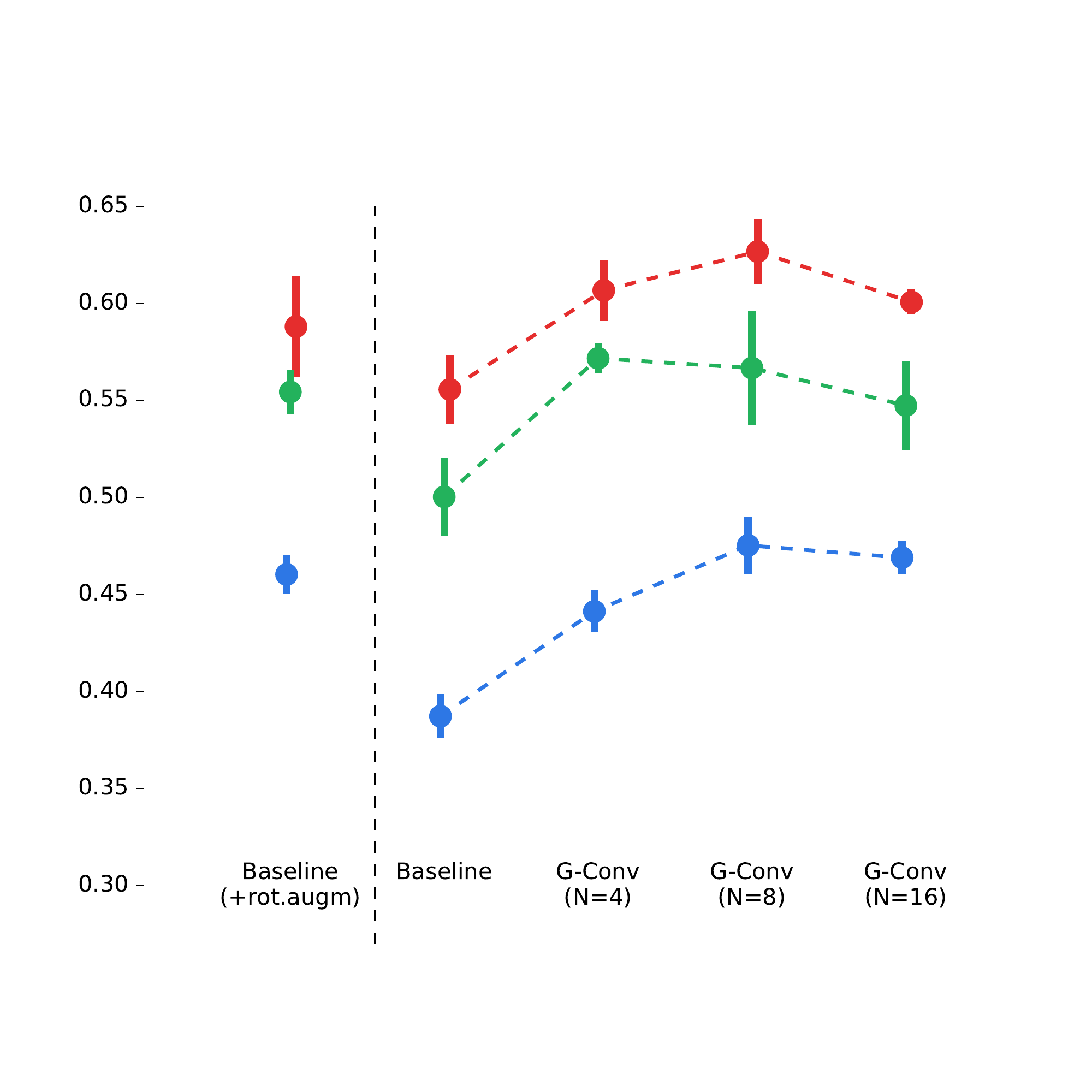}
\caption{
\footnotesize
Mean and Standard Deviation plots summarizing the \fscore{1} of the mitosis detection models.
Mean $\pm$ standard deviation is indicated.
Color identifies the different data regime (red: 8 cases; green: 4 cases; blue: 2 cases).}
\label{fig:boxPlotMitosis}
\end{center}
\end{figure}

\begin{figure}[ht!]
\begin{center}
\includegraphics[width=\columnwidth, trim=20pt 90pt 50pt 90pt, clip]{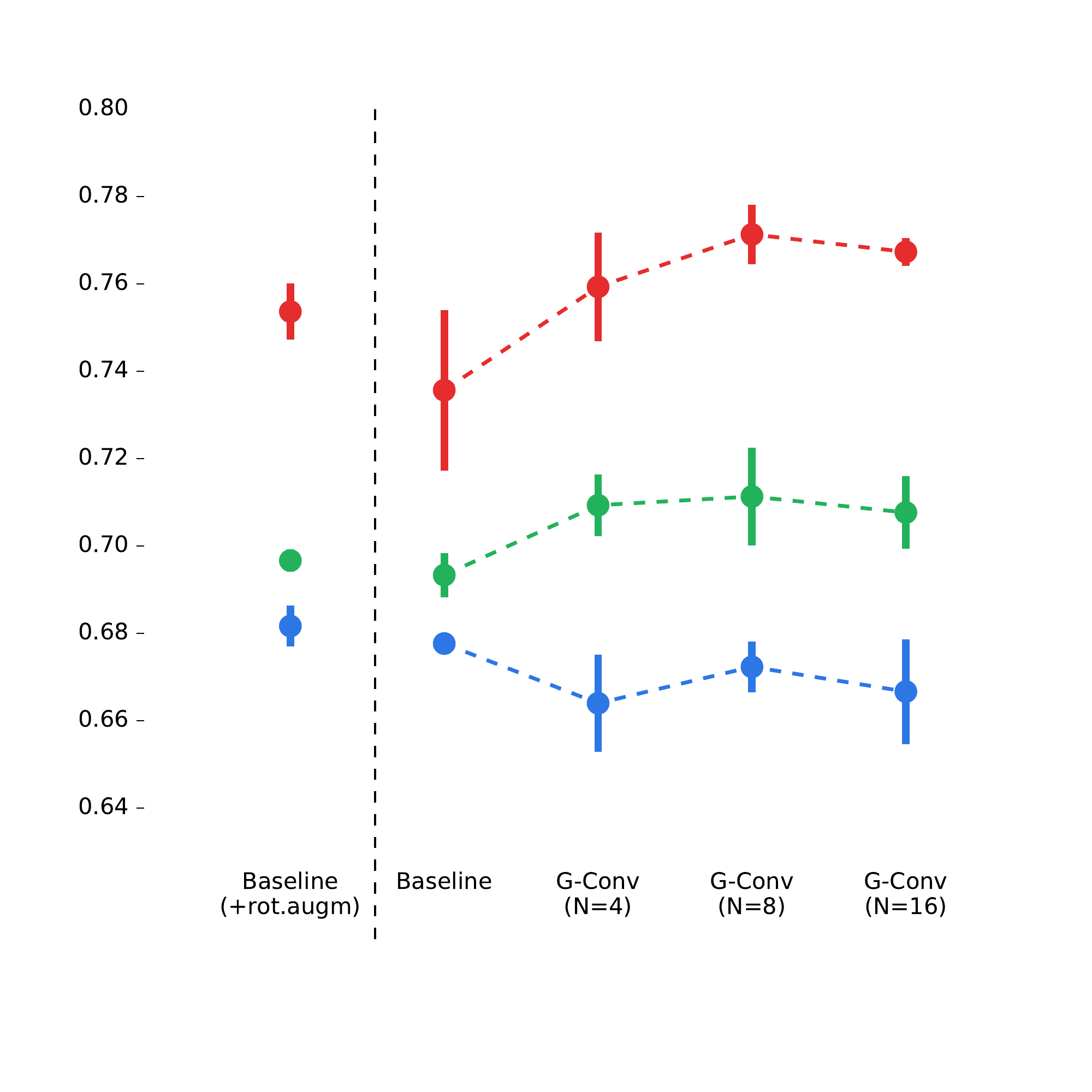}
\caption{
\footnotesize
Mean and Standard Deviation plots summarizing the \fscore{1} of the nuclei segmentation models.
Mean $\pm$ standard deviation is indicated.
Color identifies the different data regime (red: 6 HPFs/organ; green: 4 HPFs/organ; blue: 2 HPFs/organ).}
\label{fig:boxPlotNuclei}
\end{center}
\end{figure}

\begin{figure}[ht!]
\begin{center}
\includegraphics[width=\columnwidth, trim=20pt 90pt 50pt 90pt, clip]{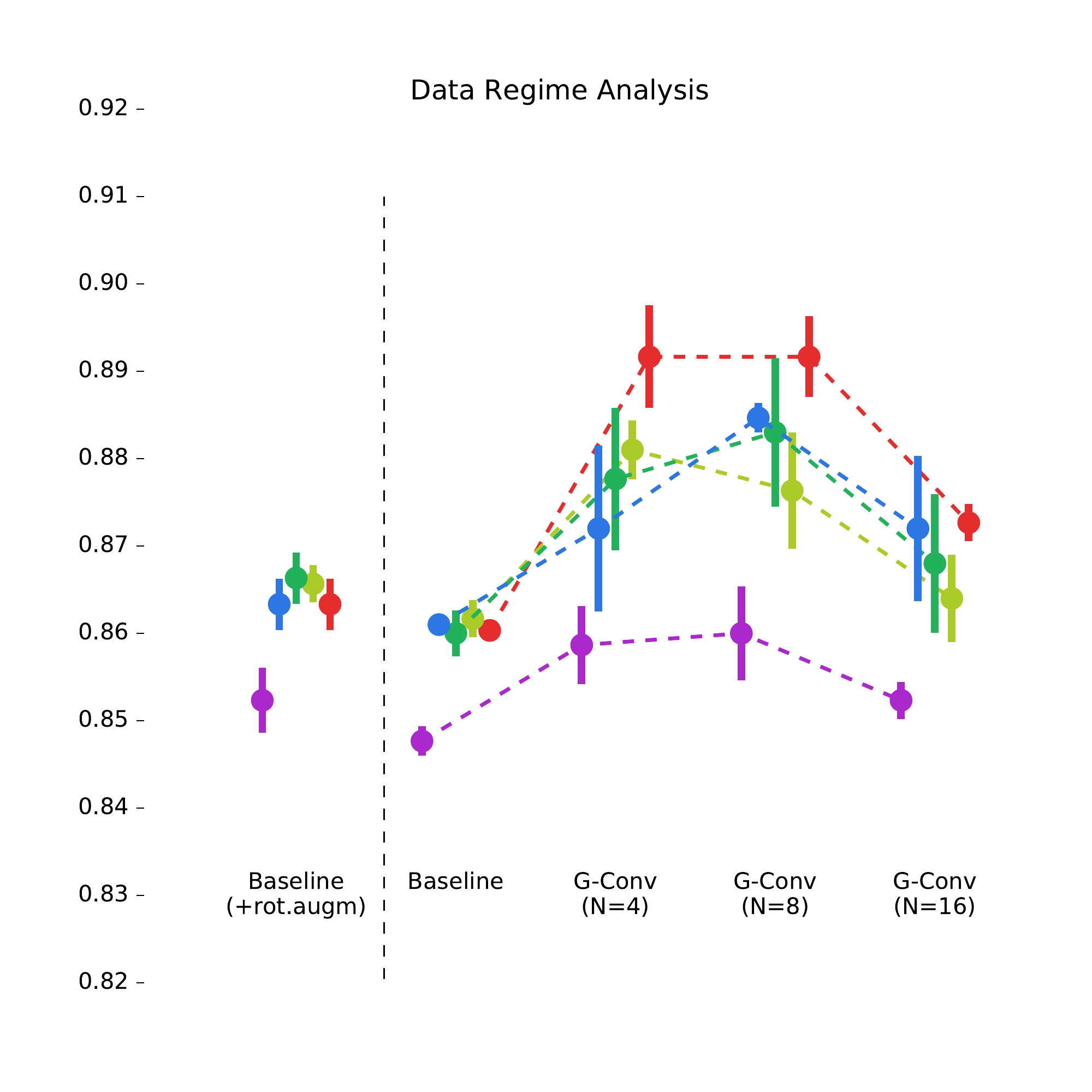}
\caption{
\footnotesize
Mean and Standard Deviation plots summarizing the accuracy of the tumor classification models.
Mean $\pm$ standard deviation  is indicated.
Color identifies the different data regime (red: 100\%; lime: 75\%; green: 50\%; blue: 25\%; purple: 10\%).}
\label{fig:boxPlotTumor}
\end{center}
\end{figure}

\subsection{Quantitative Results}
\label{quantitativeResults}
The performances of the trained models for both orientation sampling experiments and data regime experiments are summarized in the box plots of Fig. \ref{fig:boxPlotMitosis}, \ref{fig:boxPlotNuclei} and \ref{fig:boxPlotTumor}.

\paragraph{Effect of orientation sampling} 
For all three studied tasks, we observed an increase of performance with the number of sampled orientations from $N=1$ to $N=8$.
For the full data regime of the mitosis detection experiments, the use of a \se{2,8} G-CNN improves the \fscore{1} to $0.626{\pm}0.015$ on average compared to $0.556{\pm}0.016$ for the baseline model without test-time rotation augmentation (see Fig. \ref{fig:boxPlotMitosis}).
A similar increase of performances is observed for the nuclei segmentation experiments with an improvement of the \fscore{1} from $0.754{\pm}0.006$ to $0.771{\pm}0.06$ (see Fig. \ref{fig:boxPlotNuclei}), and for the tumor classification experiments with an improvement of the accuracy from $0.863{\pm}0.003$ to $0.892{\pm}0.004$ (see Fig. \ref{fig:boxPlotTumor}).

We remark that the performances of the \se{2,4} G-CNN models are better than the baseline with test-time rotation augmentation as was previously reported in literature for similar tasks \citep{bekkers2018roto, veeling2018rotation}. We also report that for all three tasks, \se{2,16} G-CNN models perform worse than the \se{2,8} G-CNN models.

\paragraph{Effect of reduced data regime with orientation sampling} 
For all three tasks, we see a global consistent decrease of performances when less training data is available.
In Fig. \ref{fig:boxPlotTumor}, the performances of the \se{2,4} and \se{2,8} G-CNN models trained with the 25\%, 50\% and 75\% data regimes, are higher than for the baseline model at full data regime using test-time rotation augmentation.
This reveals that under experimental conditions, data availability is not the only reason for limited performances since this experiment shows that the \se{2,N} G-CNN models enable achieving higher performances than the baseline models, even if less data is available.

\section{Discussion and Conclusions}
\label{discussion}
The presented study investigated the effects of embedding the \se{2} group structure in CNNs, in the context of histopathology image analysis, across multiple controlled experimental setups.

The comparative analysis we conducted shows a consistent increase of performances for three different histopathology image analysis tasks when using the proposed \se{2,N} G-CNN architecture compared to conventional CNNs acting in \mset{R}{2} evaluated with test-time rotation augmentation.
This is in line with previously reported results when using G-CNNs with groups that lay on the pixel grid (p4, p4m) \citep{cohen2016group, veeling2018rotation}, but we also show that these performances can be surpassed when using groups with higher discretization levels of \se{2}.

This confirms that conventional \mset{R}{2} CNNs struggle to learn a rotation equivariant representation based on data solely and that enforcing equivariant representation learning enables reaching higher performances.
G-CNNs with \se{2,N} structure have the advantage to guarantee higher robustness to input orientation without requiring training-time or test-time rotation augmentation.
Furthermore, the slight computational overhead for computing rotated convolutional operators and their gradient, at training time, can be canceled at test-time by computing and fixing all final oriented \se{2,N} kernels, resulting in a model that is computationally equivalent to conventional \mset{R}{2} CNNs.

We show that these performances can be surpassed when using representations with higher angular resolution levels, as shown with experiments involving \se{2,8} G-CNNs and when the training data is of sufficient amount. This conclusion corroborates the results we reported on other medical image analysis tasks \citep{bekkers2018roto} and in studies that investigated models with rotated operators that lay outside of the pixel grid \citep{hoogeboom2018hexaconv}.

However, we also identified consistent lower performances for \se{2,16} G-CNNs compared to \se{2,8} G-CNNs at full data regime.
We assume that this phenomenon is in part related to the model architectures we chose to enforce fixed model capacity, resulting in a number of channels in the representation of the \se{2,N} models being reduced when $N$ increases. This reduced number of channels might affect the diversity of the features learned by the models, to the point that this limits their overall performances.
Therefore, it appears there is a trade-off between performances and angular resolution at fixed capacity, further work would be necessary to confirm this hypothesis.

For the tumor classification task, we observed that the performances of the baseline models (with or without test-time rotation augmentation) reached a plateau, whatever the regime of available training data was among 25\%, 50\%, 75\% or 100\%.
This indicates that in the conditions of the \textit{PCam} dataset, the amount of available training data does not significantly influence the performances.
However, the rotation-equivariant models were able to achieve better performances with increased data regime.

This behavior was not evidenced for the mitosis detection and nuclei segmentation experiments.
We assume this result may be task-dependent or might be due to the fact that the plateau of performances observed for the tumor classification models was not reached yet for the two other tasks.

We qualitatively showed that in some cases, the predictions of conventional CNNs are inconsistent when inputs are rotated, whereas \se{2} G-CNNs show better stability in that sense.
This suggests that the anisotropic learned features of conventional models only get activated when the input is observed in a specific orientation.
On the shown examples (Sect. \ref{qualitativeResults}), the \se{2} models are more robust to the input orientation since their \se{2} structure guarantees the features to be expressed in multiple orientations.
We also see that \se{2} models with a limited angular resolution can yet produce some variance for rotation angles lower than this resolution.
This is also supported by the fact that higher performances were obtained for the experiments that compare \se{2,4} models to \se{2,8} models.

Still, variation of performances for these models was also observed when the input was rotated out of the pixel grid.
We explain this limit from the approximation errors caused by two of the operators we used, and that have a weaker rotation equivariance property.
First, the interpolation-based computation of the rotated kernels can cause small variations in the output when the input is rotated.
Second, the pooling operators are not rotation equivariant by construction (since they lay on fixed down-sampled versions of the pixel grid), and so are another source of error.

In conclusion, we proposed a framework for \se{2} group-convolutional network and showed its advantages for histopathology image analysis tasks.
This framework enables the learned models to be invariant to the natural roto-translational symmetry of histology images.
We showed that G-CNNs models whose representation have a \se{2} structure yield better performances than conventional CNNs and our experiments suggest the ability of G-CNNs models to fully exploit the data amount of large datasets.
Our results suggest the existence of a trade-off between network capacity and the chosen angular resolution of the \se{2,N} operators.
Directions for future work include further analysis of the relationship between the newly introduced architecture-related hyper-parameters and their effect on model performances, as well as studying other prior structures that can improve model stability to other families of input transformations.

\bibliographystyle{abbrvnat}
\balance
\bibliography{references}

\end{document}